# Constrained Cohort Intelligence using Static and Dynamic Penalty Function Approach for Mechanical Components Design

Omkar Kulkarni[1], Ninad Kulkarni[1], Anand J Kulkarni[2,3]*, Ganesh Kakandikar[1]

[1] Department of Mechanical Engineering, Faculty of Engineering, Zeal College of Engineering and Research, Pune

[2] Odette School of Business, University of Windsor, 401 Sunset Avenue, Windsor, Ontario N9B 3P4, Canada
Email: kulk0003@uwindsor.ca; Ph: 1 519 253 3000 x4939

[3]Department of Mechanical Engineering, Symbiosis Institute of Technology, Symbiosis International University, Pune MH 412 115 India
Email: anand.kulkarni@sitpune.edu.in; kulk0003@ntu.edu.sg, Ph: 91 20 39116468

**Abstract**

Most of the metaheuristics can efficiently solve unconstrained problems; however, their performance may degenerate if the constraints are involved. This paper proposes two constraint handling approaches for an emerging metaheuristic of Cohort Intelligence (CI). More specifically CI with static penalty function approach (SCI) and CI with dynamic penalty function approach (DCI) are proposed. The approaches have been tested by solving several constrained test problems. The performance of the SCI and DCI have been compared with algorithms like GA, PSO, ABC, d-Ds. In addition, as well as three real world problems from mechanical engineering domain with improved solutions. The results were satisfactory and validated the applicability of CI methodology for solving real world problems.

## 1. Introduction

In past few years, several metaheuristics have been proposed in the field of optimization. They include number of nature/bio-inspired optimization techniques such as Evolutionary Algorithms (EAs), Swarm Intelligence (SI), etc. Some of the SI techniques include Particle swarm optimization (PSO), Cuckoo Search Algorithm (CS) (Yang et al. 2009), Grey Wolf Optimizer (GWO) (Mirjalili et al. 2014), Artificial Bee Colony Algorithm (ABC) (Karaboga et al. 2011), Firefly Algorithm (FA) (Fister et.al. 2013), Ant Colony optimization (ACO) (Dorigo et al. 1997), etc. The evolutionary algorithms include Genetic Algorithm (GA) (Mitchell 1996), Differential Evolution (DE) (Storn et al. 2013; Qin et al. 2009), etc.

Cohort intelligence (CI) algorithm is a socio-inspired optimization algorithm proposed by Kulkarni, Durugkar and Kumar in 2013. The algorithm mimics the self-supervised learning behavior of cohort candidates. The cohort here refers to a group of candidates interacting and competing with one another to achieve some individual goal, which is inherently common to all. So far, CI has been applied for solving several unconstrained test problems (Kulkarni et al. 2013)**.** In addition, it was modified using a mutation approach, which expanded the search space and helped CI algorithm jump out of local minima. Also, CI and modified CI (MCI) were hybridized with K-means (Krishnasamy et al. 2014) and applied for solving several data clustering problems. The hybridized approaches exhibited exceedingly improved performance over the other contemporary algorithms such as K-means, K-means++, GA, Simulated Annealing (SA), Tabu Search (TS), ACO, Honey-bee mating optimization (HBMO) algorithm, PSO, CI and MCI. The approach of CI was also applied for solving several cases of combinatorial problems such as 0-1 Knapsack problems (Kulkarni et al. 2016). The performance was comparable to other contemporary algorithm such as HS, IHS, NGHS, QICSA and QIHSA (Kulkarni et al. 2016). The approach was also applied for solving traveling salesman problem (Kulkarni et al. 2017)**.** In addition, CI was successfully applied for solving large sized combinatorial problems with applications to healthcare, sea-cargo mix problem associated with logistics as well as cross-border supply chain problems (Kulkarni et al. 2016). It is important to mention that CI could handle the constraints using a specially developed probability based approach.

It is important to mention that the real world problems are constrained in nature. In order to make CI more generalized and powerful, constraint handling techniques need to be developed and incorporated into it. The most common approach to solve constrained optimization problems is by using a penalty functions methods. The penalty techniques transform a constrained problem into an unconstrained problem by penalizing the objective function when constraints are violated, and then minimize this penalized objective function using unconstrained optimization methods (Arora 2004). There are various penalty based constraint handling approaches developed such as static penalty approach (Homaifar et al. 1994), dynamic penalty approach (Joines & Houck 1994), annealing penalty approach (Michalewicz et al. 1994), superiority of feasible points (Powell et al. 1993), exact penalty function method (Huyer et al. 2003), barrier function methods (Arora 2004) etc. The barrier function methods are the simplest and popular method and are applicable only to the inequality constrained problems. Equality constraints may not be solvable using barrier function methods. Annealing penalty are based on the ideology of simulated annealing algorithm. This penalty approach starts with the separation of all constraints in their sub groups as linear equations, linear inequalities, nonlinear equations and nonlinear inequalities then initiating the starting point that satisfies linear constraints and then the penalty is induced (Michalewicz et al. 1994). Static penalty approach implies penalty at a constant rate whenever the constraint is violated. Dynamic penalty approach implies penalty at a dynamic or increasing rate whenever the constraint is violated. Exact penalty function method is quite promising and effective method (Ma & Zhang 2015). To avoid the difficulty of adding a large value of penalty parameter in the function the exact penalty function is used (Lucidi & Rinaldi 2010). Exact penalty can be known as function having property of recovering an exact solution of the original problem for reasonable finite values of the penalty parameter (Ma & Li 2012).

This paper investigates the constraint handling ability of CI with static penalty approach (SCI) and dynamic penalty approach (DCI). In this study, these two methods are preferred due to their advantages over other methods. The static penalty approach is characterized by its simplicity of formulation. Also, the convergence of the solutions is obtained at a faster rate as compared to other penalty function approaches. The dynamic penalty approach required small value of the penalty to be applied at the initial stages; however, as the iterations increase, the penalty also gradually increases (Mezura-Montes et al. 2011). The performance of the SCI and DCI approaches is tested by solving 20 well-known test problems from continuous domain including Pressure Vessel Design problem, Tension-Compression Spring Design problem and Welded Beam Design Problem (Liang et al. 2006; Mezura-Montes et al. 2007). The solutions are compared with several existing techniques. In addition, the Springback effect problem occurring in the automotive punch plate, thinning in connector and thickening in tail cap during the process of deep drawing are successfully solved. This validated the applicability of the proposed SCI and DCI.

The remainder of this paper is organized as follows: In Section 2, the framework of CI algorithm along with the SCI and DCI approaches are discussed. In section 3, SCI and DCI experimental evaluations and comparisons are provided. Then in section 4, application of these approaches in mechanical engineering is discussed. Finally, in section 5, conclusions and future work are provided.

## 2. Framework of Cohort Intelligence (CI) Algorithm

Cohort Intelligence (CI) algorithm models the ability of candidates in a cohort to self-supervise and improve their independent behavior. Every candidate has its own qualities which defines its behavior. In every learning attempt, each candidate tries to improve its own behavior through learning which is possible through interaction and competition with other candidates. In other words, the qualities learned eventually make every candidate improve its behavior. After certain learning attempts, the behavior of every candidate saturates and makes entire cohort converge to a unique behavior.

The Framework of the CI algorithm can be explained as follows:

**Step 1:** The qualities of the candidates are referred to as the decision variables in the problem. These qualities are randomly generated from within the associated sampling intervals.

**Step 2:** The qualities of each candidate define the behavior of the candidate. The behavior of the candidate refers to the objective function is evaluated.

**Step 3:** The probability of every candidate of being followed by other candidates is calculated on the basis of its behavior. The candidate with the best behavior has a maximum probability of being followed by other candidates and vice versa.

**Step 4:** Each candidate employs roulette wheel approach to follow a behavior in the cohort and further improves its behavior by shrinking/expanding the sampling interval of every quality of the candidate being followed. This constitutes a single learning attempt.

**Step 5:** The Algorithm is assumed to have converged on completion of the maximum number of learning attempts or the difference between the behavior of every cohort candidate is not very significant for successive considerable number of learning attempts.

**General Constrained optimization problem:**

Consider a general constrained problem (in the minimization sense) as follows:

Minimize $f(x) = f(x_1, \ldots, x_i, \ldots, x_N)$ (1)

Subject to

$g_i(x) \leq 0, \quad i = 1 \ldots n$

$h_j(x) = 0, \quad j = 1 \ldots m$

$\Psi_i^{lower} \leq x_i \leq \Psi_i^{upper}, \quad i = 1, \ldots, N.$

In the context of CI, the variables $x = (x_1, \ldots, x_i, \ldots, x_N)$ are considered as qualities. The CI optimization procedure begins with the initialization of number of candidates $C$, Sampling interval $\Psi_i$ for each quality $x_i$, $i = 1,2, \ldots, N$, learning attempt counter $l = 1$, and setting up of static sampling interval reduction factor $r \in [0,1]$, convergence parameter ε.

The penalty function approach is used to convert a constrained optimization problem into unconstrained optimization problem. Here we use two types of penalty function approaches. They are discussed below.

   a. **Static Penalty Function Method** (Homaifar et al. 1994)

A simple method to penalize infeasible solutions is to apply a constant penalty to those solutions that violate feasibility in any way. The penalty function for a problem with equality and inequality constraints can be added to form the pseudo-objective function $f_q(x)$ as follows

$$f_q(x) = f(x) + \sum_{i=1}^{n} q_i \times S \times (g_i(x))^2 + \sum_{j=1}^{m} B_j \times S \times h_j(x) \quad (2)$$

where  $f_q(x)$ is the expanded penalized objective function
   $S$ is a penalty imposed for violation of a constraint
   $q_i = 1$, if constraint $i$ is violated.
   $q_i = 0$, if constraint $i$ is satisfied.
   $B_j = 1$, if constraint $i$ is violated.
   $B_j = 0$, if constraint $i$ is satisfied.

$n$ and $m$ are the number of inequality and equality constraints, respectively. It is illustrated below with one inequality constraint.

$$\text{Minimize } f(x) = -x_1 x_2 \tag{3}$$
$$\text{Subject to } g(x) = x_1 + x_2 - 4 \leq 0$$

The associated pseudo-objective function $f_q(x)$ is as follows:

$$f_q(x) = -x_1 * x_2 + \sum_{i=1}^{n} q_i \times S \times (x_1 + x_2 - 4)^2 \tag{4}$$

If the constraint is violated the value of the $q_i$ will be *one* else, it will be *zero*. The value of penalty $S$ is chosen based on preliminary trials of the algorithm. As the learning attempts increase the violation of the constraint may decrease and eventually optimum value for the problem is achieved.

   **b. Dynamic Penalty Function** (Joines & Houck 1994)

In this constraint handling technique, the individuals are evaluated based on following formula:

$$f_q(x) = f(x) + \sum_{i=1}^{n}((q_k)^\alpha \times S) \times (g_i(x))^\beta \tag{5}$$

where $\alpha$ and $\beta$ are integer constants.

In this approach, initially a very less penalty is applied for infeasible solutions and as the algorithm progresses, penalty is increased in every learning attempt. Here $q_k$ is the iteration number which is multiplied by penalty constant $S$; however, it is very sensitive to the parameters $\alpha$ and $\beta$ and the parameters need to be properly tuned based on the preliminary trials of the algorithm. In this technique, the penalty components have a significant effect on the objective function as they increase with every learning attempt. This approach is illustrated below with one inequality constraint.

$$\text{Minimize } f(x) = -x_1 x_2 \tag{7}$$
$$\text{Subject to } g(x) = x_1 + x_2 - 4 \leq 0$$

The associated pseudo-objective function $f_q(x)$ is as follows:

$$f_q(x) = -x_1 * x_2 + \sum_{i=1}^{n}(q_k^\alpha \times S) \times (x_1 + x_2 - 4)^\beta \tag{8}$$

The value of penalty $S$ is set to be a constant; however, it is multiplied by the iteration number $q_k^\alpha$ so as the iteration increases the value of the penalty also increases. Refer to Figure 1 for CI pseudo code for both the penalty function approaches. The iterative CI procedure solving penalized function $f_q(x)$ is discussed below.

**Step 1:** The probability of selecting the behavior $f_q(x^c)$ of every associated candidate $c$ ($c = 1, \ldots, C$) is calculated as follows:

$$p^c = \frac{1/f_q(x^c)}{\sum_{c=1}^{C} 1/f_q(x^c)} \quad c\ (c = 1, \ldots, C) \tag{9}$$

**Step 2:** Every candidate $c$ generates a random number $rand \in [0,1]$ and using a roulette wheel approach decides to follow corresponding behavior $f_q(x^{c\sim})$ and associated qualities $x^{c\sim} = x_1^{c\sim}, \ldots x_i^{c\sim}, \ldots x_N^{c\sim}$. The superscript $\sim$ indicates that the behavior is selected by candidate $c$ and not known in advance. The roulette wheel approach could be most appropriate as it provides chance to every behavior in the cohort to get selected based on its quality as well as helps to incorporate uncertainty. In addition, it may increase the chances of any candidate to select the better behaviour as the associated probability stake $p^C (c = 1, ..., C)$ presented in Eq. (9) in the interval $[0, 1]$ is directly proportional to the quality of the behaviour $f_q(x^c)$. In other words, better the solution, higher is the probability of being followed by the candidates in the cohort.

**Step 3:** Every candidate $c$ $(c = 1, ..., C)$ shrinks the sampling interval $\Psi_i^{c\sim}$, $i = 1, 2, .. N$ associated with every variable $x_i^{c\sim}$, $i = 1, 2, .. N$ to its local neighborhood. This is done as follows:

$$\Psi_i^{c\sim} = [x_i^{c\sim} - \frac{\|\Psi_i^{upper} - \Psi_i^{lower}\|}{2} \times r \, , \, x_i^{c\sim} + \frac{\|\Psi_i^{upper} - \Psi_i^{lower}\|}{2} \times r ] \tag{10}$$

**Step 4:** Every candidate $c$ $(c = 1, ..., C)$ samples $t$ qualities from within the updated sampling interval $\Psi_i^{c\sim}$, $i = 1, 2, .. N$ associated with every quality $x_i^{c\sim}$, $i = 1, 2, .. N$ and computes a set of associated $t$ behaviors, i.e. $F_q^{c,t} = \{f_q(x^c)^1, ..., f_q(x^c)^j, ..., f_q(x^c)^t\}$ and selects the best behavior $f_q^*(x^c)$ from within. This makes the cohort available with $C$ updated behaviors represented as $F_q^{*C} = \{f^*(x^1), ... f^*(x^c), ... f^*(x^C)\}$.

**Step 5:** The cohort behavior could be considered saturated, if there is no significant improvement in the behavior $f^*(x^c)$ of every candidate $c$ $(c = 1, ..., C)$ in the cohort, and the difference between the individual behaviors is not very significant for successive considerable number of learning attempts, i.e. if

1. $\left\| \max(F_q^{*C})^n - \max(F_q^{*C})^{n-1} \right\| \leq \epsilon$, and
2. $\left\| \min(F_q^{*C})^n - \min(F_q^{*C})^{n-1} \right\| \leq \epsilon$, and
3. $\left\| \max(F_q^{*C})^n - \min(F_q^{*C})^n \right\| \leq \epsilon$, every candidate $c$ $(c = 1, ..., C)$ expands the sampling interval $\Psi_i^{c\sim}$, $i = 1, ..., N$ associated with every quality $x_i^{c\sim}$, $i = 1, .., N$ to its original one $\Psi_i^{lower} \leq x_i \leq \Psi_i^{upper}$, $i = 1 ..., N$.

**Step 6:** If either of the two criteria listed below is valid, accept any of the $C$ behaviors from current set of behaviors in the cohort as the final objective function value $f^*(x)$ as the final solution and stop, else continue to Step 1.
   a) If maximum number of learning attempts exceeded.
   b) If cohort saturates to the same behavior (satisfying the conditions in Step 5) for significant number of times.

```
Objective function f(x), x = (x_1,...,x_N)
Inequality Constraints g_i(x) < 0, i = 1,...,n
Equality Constraints h_j(x) = 0, j = 1,...,m
Select number of candidates in Cohort (C).
Set sampling interval reduction factor (r).
Set convergence factor (ε),
While (t < Max Learning attempts) or Convergence criterion
        Initialize qualities/attributes of each candidate by random numbers
        Evaluate the values constraints for each candidate
                If (Constraints Satisfied)
                        Do not apply penalty
                Else
                        Apply penalty function approach (Static or Dynamic) *
                End
        Evaluate fitness/behavior of every candidate.
        Evaluate probability associated with the behavior being followed by every
        candidate in the cohort by using equation
        Use roulette wheel approach to select behavior to follow by each candidate within
        Cohort (C).
        Every candidate will shrinks the sampling interval of every quality with respect to
        the followed behavior
        If (Convergence criterion met)
        Accept the current best candidate and its behavior
                Break
        End
End
```

**Fig. 1. Pseudo code of CI algorithm**

## 3. Experimental Evaluations

Performance of SCI and DCI was tested by evaluation of 20 well known test problems (Liang J. J. et al. 2006, Mezura-Montes et al. 1994). The characteristics of these problems are listed in Table 1. In addition to these test functions, mechanical engineering problems such as like tension compression string, welded beam design and Pressure vessel design were also solved. The SCI and DCI were coded in Matlab (R2014a) on windows 7 platform with I5-3470 Processor 3.2GHZ processor speed and 4GB RAM. Every problem was solved 20 times with the CI parameters chosen as follows: number of candidates $C = 5$, reduction factor $r = 0.9$ and convergence factor $\varepsilon = 1E - 11$. These parameters were chosen based on the preliminary trials of the algorithm.

**Table 1. Features of Test Problems**

| Problem | Dimension | Type | LI* | NI* | LE* | NE* |
|---|---|---|---|---|---|---|
| G01 | 13 | Quadratic | 9 | 0 | 0 | 0 |
| G02 | 20 | Non-linear | 0 | 2 | 0 | 0 |
| G03 | 10 | Polynomial | 0 | 0 | 0 | 1 |
| G04 | 5 | Quadratic | 0 | 6 | 0 | 0 |

| Problem | n | Type | LI | NI | LE | NE |
|---|---|---|---|---|---|---|
| G05 | 4 | Cubic | 2 | 0 | 0 | 3 |
| G06 | 2 | Cubic | 0 | 2 | 0 | 0 |
| G07 | 10 | Quadratic | 3 | 5 | 0 | 0 |
| G08 | 2 | Non-linear | 0 | 2 | 0 | 0 |
| G09 | 7 | Polynomial | 0 | 4 | 0 | 0 |
| G10 | 8 | Linear | 3 | 3 | 0 | 0 |
| G11 | 2 | Quadratic | 0 | 0 | 0 | 1 |
| G12 | 3 | Quadratic | 0 | 1 | 0 | 0 |
| G14 | 10 | Non-linear | 0 | 0 | 3 | 0 |
| G15 | 3 | Quadratic | 0 | 0 | 1 | 1 |
| G17 | 6 | Non-linear | 0 | 0 | 0 | 4 |
| G18 | 9 | Quadratic | 0 | 13 | 0 | 0 |
| G24 | 2 | Linear | 0 | 2 | 0 | 0 |

*LI: linear inequality; NI: nonlinear inequality; LE: linear equality; NE: nonlinear equality

The performance of SCI & DCI is compared with the other methods in relevant literature including GA (Mezura-Montes and Coello 2005), PSO (Zavala et al. 2005), ABC (Karaboga et al. 2011) & Dynamic-Differential Search algorithm (D-DS) (Jianjun et al. 2015, Mezura-Montes et al. 2007). These selected algorithms have their special qualities which led them in reaching the optimized solutions of the constrained test problems. GA is an optimization algorithm which is based on the Darwin's theory of evolution. While solving constrained test problems using GA, Deb's feasibility rules were used as a constraint handling technique which led them overcome infeasible solutions. Similarly, PSO is inspired by social behavior of bird flocking or fish schooling. PSO also used Deb's feasibility rules for handing constraints. ABC also solved the constrained optimization problems by using static penalty method as a constraint handling technique and this technique was found to be more compatible with the algorithm as it converged quicker and gave optimized solutions. Similarly, D-DS algorithm used dynamic penalty method as a constraint handling technique and the technique was found to be quite effective as the penalty was steadily increasing with increasing number of iterations. Experimental results of CI algorithm over 20 runs are provided in Table 2.

**Table 2. CI Performance of SCI and DCI**

| Problem | Optimum | Best Solution by SCI | SCI (Standard Deviation) | Best Solution by DCI | DCI (Standard Deviation) |
|---|---|---|---|---|---|
| G01 | $-15.000$ | $-14.9974$ | $0.198195$ | $-15.0000$ | $0.1473$ |
| G02 | $-0.803619$ | $-0.80357$ | $0.036091$ | $-0.8036$ | $0.0253$ |
| G03 | $1.000$ | $-1.00125$ | $0.001112$ | $-0.9997$ | $0.0008$ |
| G04 | $-30665.539$ | $-30665.5$ | $0.044975$ | $-30665.5465$ | $16.1751$ |
| G05 | $5126.498$ | $5119.139$ | $40.41773$ | $5121.1504$ | $42.1847$ |
| G06 | $-6961.814$ | $-6961.81$ | $1.52E-05$ | $-6961.8139$ | $3.83E-07$ |
| G07 | 24.306 | 24.30437 | 0.221551 | 24.3506 | 0.2135 |
| G08 | 0.095825 | −0.09583 | $1.06E-12$ | −0.0958 | $6.23E-08$ |
| G09 | 680.63 | 680.6726 | 0.259831 | 680.6738 | 0.2882 |
| G10 | 7049.25 | 7051.823 | 11.55862 | 7051.9256 | 15.3881 |
| G11 | 0.75 | 0.74965 | 0.001311 | 0.7489 | 0.0011 |
| G12 | 1.000 | −1 | $1.63E-12$ | −1.0000 | 0.0025 |
| G14 | −47.7649 | −47.735 | 0.20067 | −47.7385 | 0.1630 |
| G15 | 961.7150 | 961.7152 | 0.009111 | 961.6403 | 0.2286 |
| G18 | −0.8660 | −0.86603 | 0.001221 | −0.8660 | 0.0005 |
| G24 | −5.5080 | −5.50801 | $2.3E-07$ | −5.5080 | $2.94E-07$ |
| PV* | 6059.86326 | 5891.588 | 37.96514 | 5890.5657 | 40.4247 |
| TC* | 0.0127048 | 0.012666 | 0.000453 | 0.0127 | 0.0001 |
| WBD* | 1.748 | 2.221102 | 0.166765 | 2.2556 | 0.1044 |

*PV = Pressure Vessel Design, TC = Tension Compression Spring Design, WBD = Welded Beam Design.

As exhibited in Table 2, SCI and DCI have found solutions in the close neighborhood of the reported optimum solution for most of the problems. The standard deviation (SD) for SCI and DCI was observed to be varying with the variation in the type of problem. It shows that the constraint handling techniques used to solve the problems with equality constraints is not so compatible with CI. Table 3 shows the best solutions by SCI and DCI as well as some other methods. The results show that for problems which have comparatively larger feasible space (refer to Table 1) both SCI and DCI yielded comparative/superior results as compared to other algorithms.

Table 3 Comparison of SCI & DCI with other algorithms

| Problem | Optimum | GA | PSO | ABC | d-DS | **SCI** | **DCI** |
|---|---|---|---|---|---|---|---|
| G01 | −15.000 | −15.000 | −14.710 | −15.000 | −15.000 | **−14.997** | **−14.998** |
| G02 | −0.804 | −0.785 | −0.420 | −0.792 | −0.804 | **−0.804** | **−0.804** |
| G03 | 1.000 | 1.000 | 0.765 | 1.000 | NA | **1.001** | **1.000** |
| G04 | −30665.53 | −30665.53 | −30665.53 | −30665.53 | −30665.53 | **−30665.50** | **−30665.50** |
| G05 | 5126.498 | 5174.492 | 5135.973 | 5185.714 | 5131.343 | **5119.139** | **5121.150** |
| G06 | −6961.814 | −6961.284 | −6961.814 | −6961.813 | 6961.814 | **−6961.810** | **−6961.810** |
| G07 | 24.306 | 24.475 | 32.407 | 24.473 | 24.315 | **24.304** | **24.351** |
| G08 | 0.096 | 0.096 | 0.096 | 0.096 | 0.096 | **0.096** | **0.096** |
| G09 | 680.630 | 680.643 | 680.630 | 680.640 | 680.630 | **680.673** | **680.674** |
| G10 | 7049.250 | 7253.047 | 7205.500 | 7224.407 | 7056.760 | **7051.823** | **7051.926** |
| G11 | 0.750 | 0.750 | 0.749 | 0.750 | 0.750 | **0.750** | **0.749** |
| G12 | 1.000 | 1.000 | 0.999 | 1.000 | 1.000 | **1.000** | **1.000** |

| Problem | | | | | | |
|---|---|---|---|---|---|---|
| G14 | 47.765 | *NA* | *NA* | *NA* | 47.458 | **47.735** | 47.739 |
| G15 | 961.715 | *NA* | *NA* | *NA* | 961.715 | **961.715** | 961.640 |
| G17 | 8853.54 | *NA* | *NA* | *NA* | 8853.830 | **DNC** | 8913.786 |
| G18 | −0.866 | −0.852 | *NA* | *NA* | −0.866 | **−0.866** | −0.866 |
| G24 | −5.508 | −5.508 | *NA* | −5.507 | −5.508 | **−5.508** | −5.508 |
| PV | 6059.863 | *NA* | 6059.714 | 6059.714 | *NA* | **5891.588** | 5890.566 |
| TC | 0.013 | *NA* | 0.013 | 0.013 | *NA* | **0.013** | 0.013 |
| WBD | 1.748 | *NA* | 1.725 | 1.725 | *NA* | **2.221** | 2.256 |

**\*NA = Not Available, DNC = Did Not Converge**

It could be observed that SCI and DCI perform better than GA, PSO, D-DS and ABC algorithms. It is important to mention here that all these problems have inequality type of constraints. Table 4 shows the number of average Function Evaluations (FE) and average time required by SCI and DCI. The table exhibited that with problems with fewer constraints (G08 and G12) and fewer dimensions (G11) fewer average FE and time was required; however, with increasing number of constraints and the dimensions average FE and the computational time increased.

**Table 4 Function Evaluations & Time Required**

| Problem | FE by SCI | Time (sec) required for SCI | FE by DCI | Time (sec) required for DCI |
|---|---|---|---|---|
| G01 | 5330 | 23.54 | 5210 | 38.69 |
| G02 | 7190 | 40.97 | 7100 | 82.90 |
| G03 | 1620 | 14.38 | 1220 | 13.29 |
| G04 | 4365 | 11.76 | 1290 | 4.099 |
| G05 | 4590 | 26.37 | 4145 | 35.22 |
| G06 | 2525 | 5.82 | 2185 | 3.47 |
| G07 | 5175 | 8.70 | 4185 | 7.02 |
| G08 | 1170 | 2.44 | 625 | 1.02 |
| G09 | 2800 | 8.11 | 2735 | 7.21 |
| G10 | 17980 | 57.24 | 13675 | 41.31 |
| G11 | 1240 | 3.12 | 1275 | 5.29 |
| G12 | 950 | 5.72 | 420 | 8.35 |
| G14 | 4920 | 25.55 | 12375 | 98.52 |
| G15 | 2550 | 5.07 | 12195 | 37.83 |
| G17 | DNC* | DNC* | 6475 | 23.87 |
| G18 | 1620 | 6.59 | 3210 | 18.38 |
| G24 | 1275 | 2.6 | 4345 | 13.65 |
| PV | 7670 | 29.377 | 7455 | 27.4 |
| TC | 3590 | 12.18 | 5235 | 18.85 |
| WBD | 4865 | 13.41 | 4945 | 17.72 |

**DNC -Did Not Converge.**

As an illustration of working mechanism of CI, the convergence plots for SCI and DCI for solving problem G24 is shown in Figure 2(a) and 2(b). These plots indicate that initially, the behavior of the candidates in a cohort is different from one another; however, with increasing learning attempts, the candidates learn by following the behavior of one another, which further led to saturation of the solutions. After the first saturation, the sampling interval associated with every quality/variable is expanded in order to avoid premature convergence and hence find the global optimum. In most of the test problems, the solution obtained on every saturation were quite closer except G04 and G07.

1. **G24 Problem**

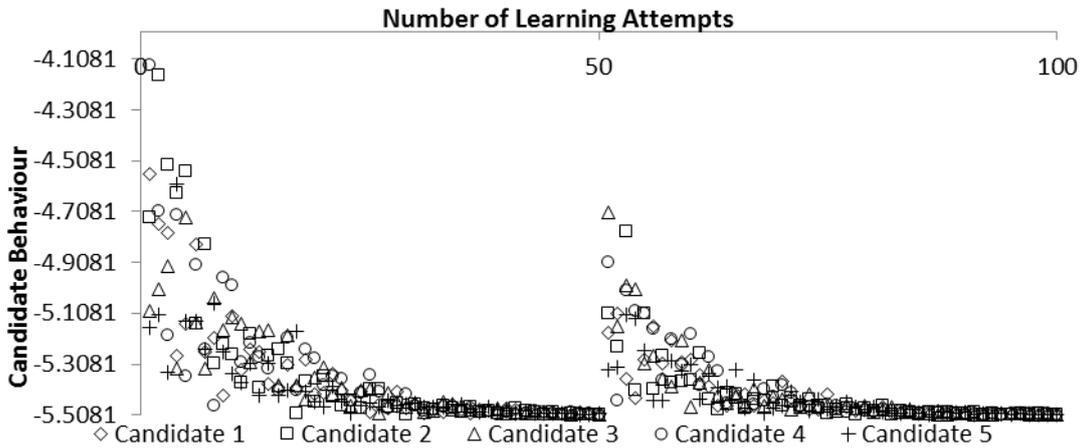

**Figure 2(a). SCI-G24**

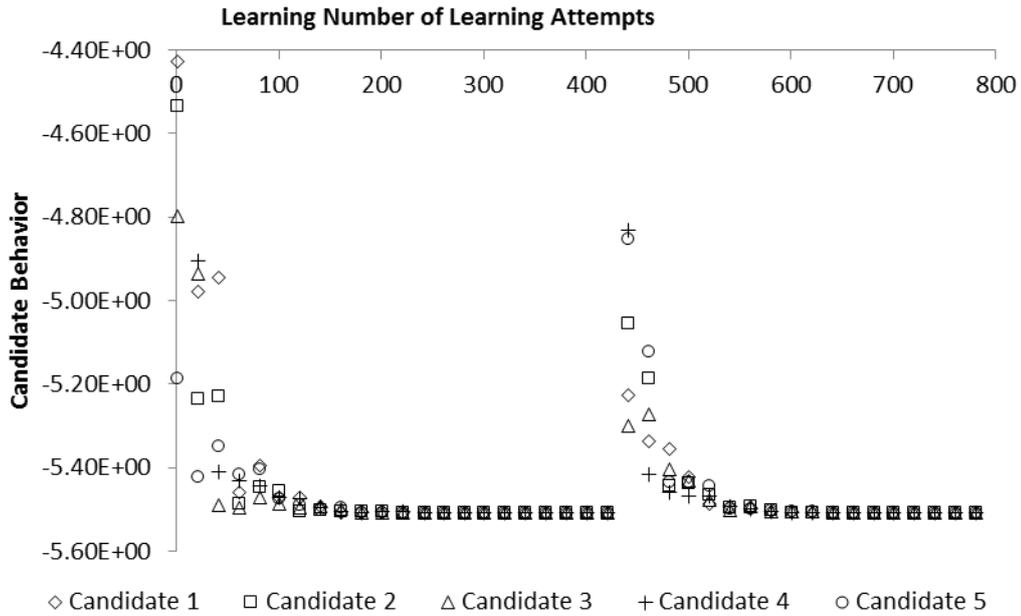

**Figure 2(b). DCI-G24**

### 4. Application of Cohort Intelligence

**Nomenclature**

| | |
|---|---|
| $BHF$ | Blank Holder Force ($N$) |
| $R_D$ | Radius on Die ($mm$) |
| $R_P$ | Radius on Punch ($mm$) |
| $\mu$ | Coefficient of Friction |
| $d_0$ | Blank Diameter ($mm$) |
| $d_1$ | Finished Component Diameter ($mm$) |
| $P$ | Pressure Applied ($N/mm^2$) |
| $S_0$ | Thickness of the Sheet ($mm$) |
| $SDM$ | Springback Displacement Magnitude |
| $z$ | Corner Radius ($mm$) |

In addition to the mechanical engineering design problems such as Pressure Vessel Design problem, Tension-Compression Spring Design problem and Welded Beam Design Problem, SCI and DCI were successfully applied to solve the Springback effect problem occurring in the automotive punch plate, thinning in connector and thickening in tail cap during the process of deep drawing (Kakandikar 2014). Similar to the above test problems the SCI and DCI were coded in MATLAB (R2014a) on windows 7 platform with intel I5-3470 Processor 3.2GHZ processor speed and 4GB RAM. Similar to the previous problems, every problem was solved 20 times with the CI parameters chosen as follows: number of candidates $C = 5$, reduction factor $r = 0.9$ and convergence factor $\varepsilon = 1E - 11$. These parameters were chosen based on the preliminary trials of the algorithm.

### 4.1. Springback Problem in Punch Plate:

Springback is the elastic recovery of the component after the mechanical drawing process is completed. This occurs in all components where the elastic property of the material is present. Springback in any component is a defect which may vary the components dimensions from the desired one. So in any component it should be minimized. The Springback optimization problem of an automotive punch plate in the process of deep drawing is solved by SCI and DCI. The problem is linear with four variables and two inequality constraints.

#### 4.1.1. Component Description

The component used is Punch Plate for the optimization process. The weight of the component is $90\ gms$. The material used for the component is SPCC steel which is a commercial quality cold rolled steel. Thickness of the material used for the process is $0.8\ mm$. Yield strength of the material used is $280\ MPa$. The ultimate tensile strength of the material is $340\ MPa$.

$Minimize$

$$SDM = 0.0488 - 0.000133 \times BHF - 0.0167 \times \mu + 0.00150 \times R_D + 0.00217 \times R_P \qquad (11)$$

Subject to
$2.5 < R_D < 8$

$3 \times R_D > R_p > 6 \times R_D$

where

$$BHF = \frac{\pi}{4}(d_o{}^2 + 2z)^2 \times P \qquad (12)$$

where $P = 2.5\ N/mm^2$.

and

$$R_D = 0.035\ [50 + (d_0 - d_1)]\sqrt{S_0} \qquad (13)$$

$R_P = (3\ to\ 6) \times R_D$

### 4.1.2. Results obtained

**Table 5 (a) Results Obtained for Original Component** (Kakandikar 2014)

| Original Component | |
|---|---|
| SDM | $0.07420\ mm$ |
| Radius on Die ($R_D$) | $2.886\ mm$ |
| Blank Holder Force ($BHF$) | $16.931\ KN$ |
| Radius on Punch ($R_P$) | $14.38111\ mm$ |
| Coefficient of Friction ($\mu$) | $0.15$ |

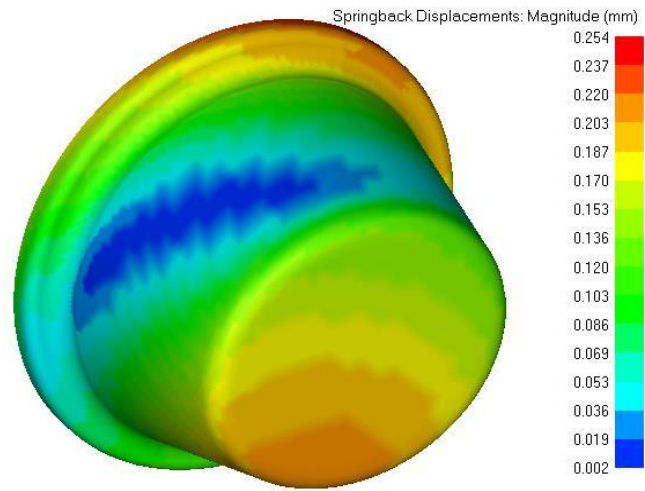

**Figure 3 (a). Springback displacement magnitude of the original component.**

**Table 5 (b) Results by SCI**

| Modified Component by SCI | |
|---|---|
| SDM | $0.06698\ mm$ |
| Radius on Die ($R_D$) | $2.8490\ mm$ |
| Blank Holder Force ($BHF$) | $16.961\ KN$ |
| Radius on Punch ($R_P$) | $8.5474\ mm$ |
| Coefficient of Friction ($\mu$) | $0.1449$ |

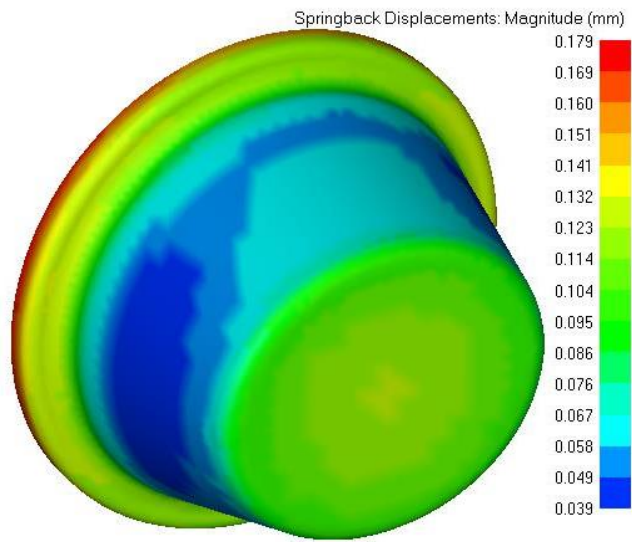

**Figure 3 (b). Springback displacement magnitude of modified component by SCI**

Table 5 (c) Results by DCI

| Modified Component By DCI | |
|---|---|
| SDM | $0.06466\ mm$ |
| Radius on Die ($R_D$) | $2.858\ mm$ |
| Blank Holder Force ($BHF$) | $17.42\ KN$ |
| Radius on Punch ($R_P$) | $8.6049\ mm$ |
| Coefficient of Friction ($\mu$) | $0.14$ |

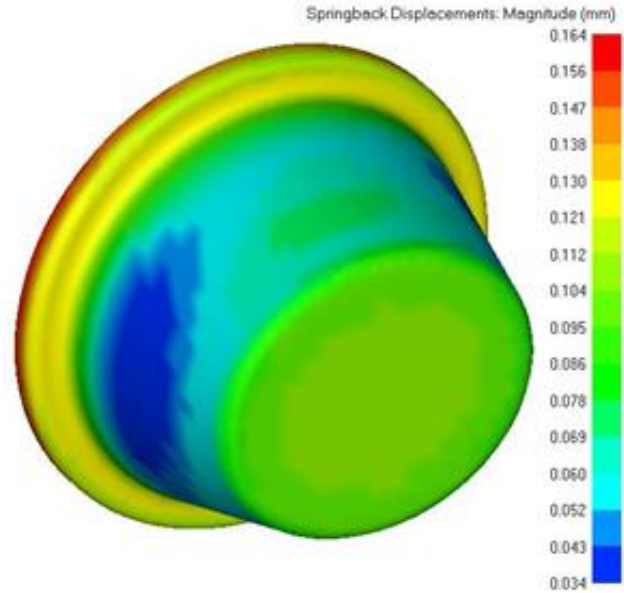

Figure 3 (c). Springback displacement magnitude of modified component by DCI

The original component design and associated formulation for the Springback displacement magnitude (SDM) is taken from (Kakandikar 2014). The problem (Eq 11 to 13) is solved using SCI and DCI. The SDM depends on blank holder force, coefficient of friction, radius on die and radius on punch. All these parameters finally depend on variables blank diameter $d_o$, finished component diameter $d_1$ and corner radius $z$. These parameters define the design of the component. The solutions using the SCI and DCI were then used to modify the original component design.

It can be observed from the Table 5(a) that the SDM obtained in the original component is $0.07420\ mm.$ The corresponding formability analysis solution performed in FromingSuite version 2015.1.0 software in Figure 3(a). It is observed that the range of the SDM for the component varied from $0.002\ to\ 0.254\ mm.$ The SDM obtained by SCI is $0.06698\ mm$ (refer to Table 5(b)) which indicates that the springback is reduced by $9.73\%$. Also it can be observed from Figure 3(b) that the reduced component SDM ranges from $0.039\ to\ 0.179\ mm$ which indicates that the average springback through overall component is reduced. Similarly the SDM obtained by DCI is $0.06466\ mm$ (refer to Table 5(c)) indicates that the springback is reduced by $12.85\%$. Also it can be observed from Figure 3(c) that the SDM for the entire component ranges from $0.034\ to\ 0.164\ mm$ this indicates that the average springback of overall component is reduced.

### 4.2. Problem of Thinning in Connector

Thinning is the most common defect occurring in the components manufactured by deep drawing process. It is necessary to minimize thinning in order to maintain the quality of the product and further may reduce the production cost of the material and time. The final objective of deep drawing process in particular or of any sheet metal forming process in general is to produce good quality product, hence uniform thickness should be obtained throughout. The thinning minimization of connector in the process of deep drawing is solved by SCI and DCI.

### 4.2.1. Component Description

The weight of the original component was 20 $gms$. The thickness of the sheet was selected as 1 $mm$. The material used was D 513, SS 4010. The Yield strength of the material was 280 $MPa$ while the ultimate tensile strength was 360 $Mpa$.

$Minimize$

$$Thinning = 1.35 - 0.0400 \times BHF - 0.733 \times \mu - 0.0300 \times R_D - 0.0183 \times R_P \quad (14)$$

Subject to

$$2 < R_D < 4$$

$$3 \times R_D > R_p > 6 \times R_D$$

where

$$BHF = \frac{\pi}{4}(d_o^2 + 2z)^2 \times P \quad (15)$$

where $P = 2.5 \; N/mm^2$

and

$$R_D = 0.035\,[50 + (d_0 - d_1)]\sqrt{S_0} \quad (16)$$

$$R_P = (3 \; to \; 6) \times R_D$$

### 4.2.2. Results Obtained

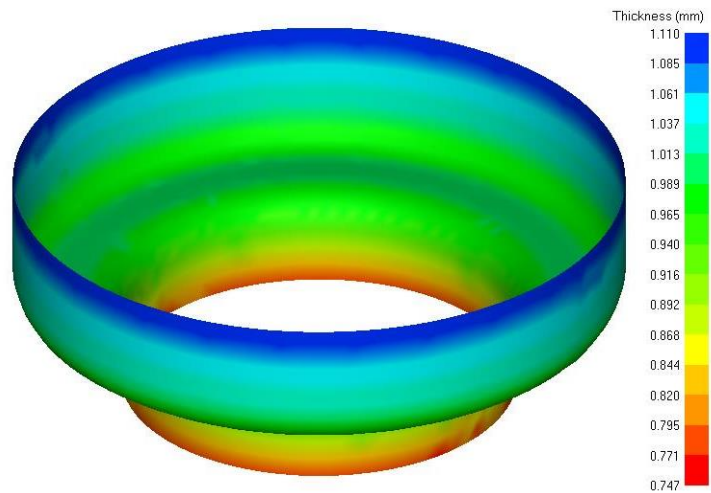

**Figure 4 (a) Thickness distribution of the original component.**

**Table 6 (a) Results Obtained for Original Component** (Kakandikar 2014)

| Original Component | |
|---|---|
| Thinning | 0.896 $mm$ |
| Radius on Die ($R_D$) | 2.52 $mm$ |
| Blank Holder Force ($BHF$) | 3.89 $KN$ |
| Radius on Punch ($R_P$) | 6.16 $mm$ |
| Coefficient of Friction (µ) | 0.15 |

**Table 6 (b) Results by SCI**

| Modified Component by SCI | |
|---|---|
| Thinning | $0.943\ mm$ |
| Radius on Die ($R_D$) | $2.50\ mm$ |
| Blank Holder Force ($BHF$) | $4.01\ KN$ |
| Radius on Punch ($R_P$) | $9.17\ mm$ |
| Coefficient of Friction ($\mu$) | $0.005$ |

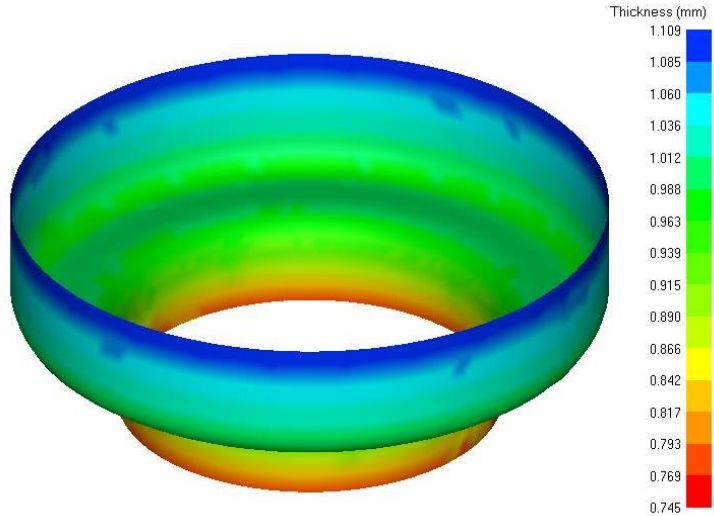

Figure 4 (b) Thickness distribution of the modified component by SCI.

**Table 6 (c) Results by DCI**

| Modified Component by DCI | |
|---|---|
| Thinning | $0.969\ mm$ |
| Radius on Die ($R_D$) | $2.50\ mm$ |
| Blank Holder Force ($BHF$) | $4.25\ KN$ |
| Radius on Punch ($R_P$) | $9.17\ mm$ |
| Coefficient of Friction ($\mu$) | $0.005$ |

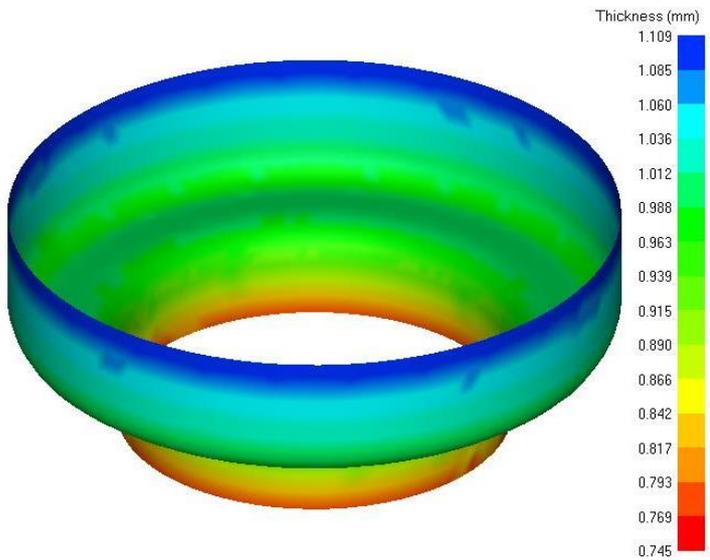

Figure 4 (c) Thickness distribution of the modified component by DCI

The original component design and mathematical formulation (Eq 14 to 16) for Thinning is taken from Kakandikar 2014. The $Thinning$ is solved using SCI and DCI. The $Thinning$ (refer to Eq 14) depends on blank holder force, coefficient of friction, radius on die and radius on punch. All these parameters finally depend on variables blank diameter $d_o$, finished component diameter $d_1$ and corner radius $z$. These parameters define the design of the component. The solutions using the SCI and DCI are then used to modify the original component design.

It can be observed from Table 6(a) that the $Thinning$ value obtained in the original component is $0.896\ mm$. The corresponding formability analysis solution performed in FromingSuite version 2015.1.0 software is shown in Figure 4(a). It is observed that the range of the thickness distribution in the overall component varies from $0.747\ to\ 1.110\ mm$. The thickness obtained from solving by SCI is $0.943\ mm$ (refer to Table 6(b)) indicates that the thickness is increased by 5.24%. also it can be observed from Figure 4(b) that the overall component thickness distribution range is reduced from $0.745\ to\ 1.109\ mm$ which indicates that the overall thickness is increased. Similarly, the thickness obtained by DCI is $0.969\ mm$ as shown in Table 6(c) this indicates that the thickness is increased by 8.45%. Also it can be observed from Figure 4(c) that the whole component thickness distribution ranges from $0.745\ to\ 1.109\ mm$. It indicates that the average thickness in overall component is increased.

### 4.3. Thickening Problem in Tail Cap

Thickening is one of the major defect occurring in the components manufactured by deep drawing process. It is necessary to minimize thickening in order to maintain the quality of the product. Determination of the thickness distribution and of the thinning of the sheet metal blank reduces the production cost of the material and time. The final objective of deep drawing process in particular or of any sheet metal forming process in general is to produce good quality product, hence uniform thickness should be obtained throughout. The thickening minimization of tail cap in the process of deep drawing was solved by SCI and DCI.

#### 4.3.1. Component Description

The weight of the original component was found to be 20 grams. The thickness of the sheet selected is $1.2\ mm$. The material used is D 513, SS 4010. The Yield strength of the material is $250\ MPa$ while the ultimate tensile strength was $350\ Mpa$.

$$Thickening = 1.278 + 0.00180 \times BHF + 0.043 \times \mu - 0.0167 \times R_D - 0.0000 \times R_P \qquad (17)$$

Subject to

$$2 < R_D < 4$$

$$3 \times R_D > R_p > 6 \times R_D$$

where

$$BHF = \frac{\pi}{4}(d_o^2 + 2z)^2 \times P \qquad (18)$$

where $P = 2.5\ N/mm^2$

and

$$R_D = 0.035\ [50 + (d_0 - d_1)\sqrt{S_0} \qquad (19)$$

$$R_P = (3\ to\ 6) \times R_D$$

### 4.3.2. Results Obtained

**Table 7 (a) Results Obtained for Original Component** (Kakandikar 2014)

| Original Component | |
|---|---|
| Thickness | $1.309\ mm$ |
| Radius on Die ($R_D$) | $3.83\ mm$ |
| Blank Holder Force ($BHF$) | $21.99\ KN$ |
| Radius on Punch ($R_P$) | $17.1\ mm$ |
| Coefficient of Friction ($\mu$) | $0.15$ |

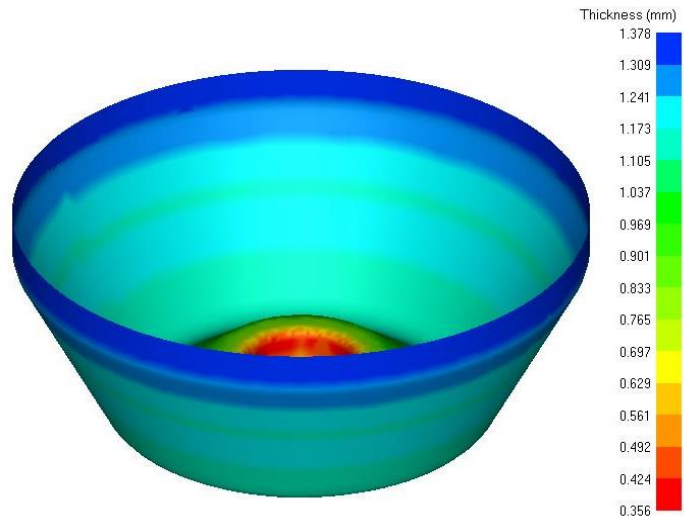

**Figure 5 (a) Thickness distribution of the original component.**

**Table 7 (b) Results by SCI**

| Modified Component by SCI | |
|---|---|
| Thickness | $1.276\ mm$ |
| Radius on Die ($R_D$) | $3.9\ mm$ |
| Blank Holder Force ($BHF$) | $23.67\ KN$ |
| Radius on Punch ($R_P$) | $17.3\ mm$ |
| Coefficient of Friction ($\mu$) | $0.005$ |

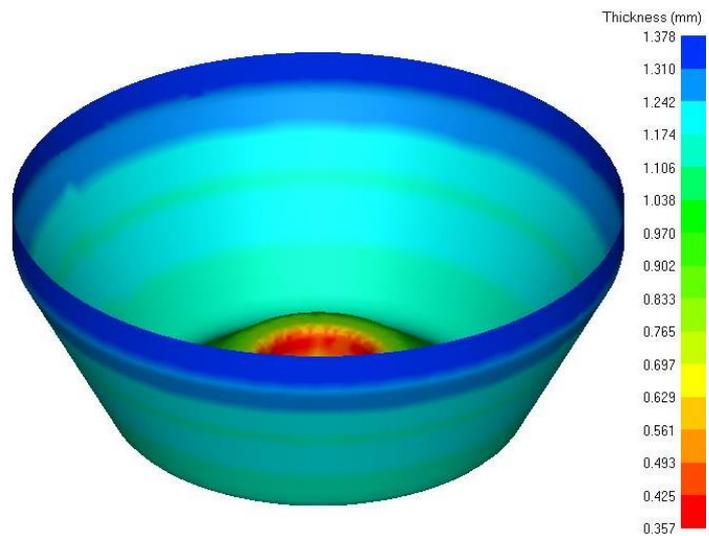

**Figure 5 (b) Thickness distribution of the modified component by SCI.**

**Table 7 (c) Results by DCI**

| Modified Component | |
|---|---|
| Thickness | $1.268\ mm$ |
| Radius on Die ($R_D$) | $3.85\ mm$ |
| Blank Holder Force ($BHF$) | $23.23\ KN$ |
| Radius on Punch ($R_P$) | $17.3\ mm$ |
| Coefficient of Friction (μ) | $0.005$ |

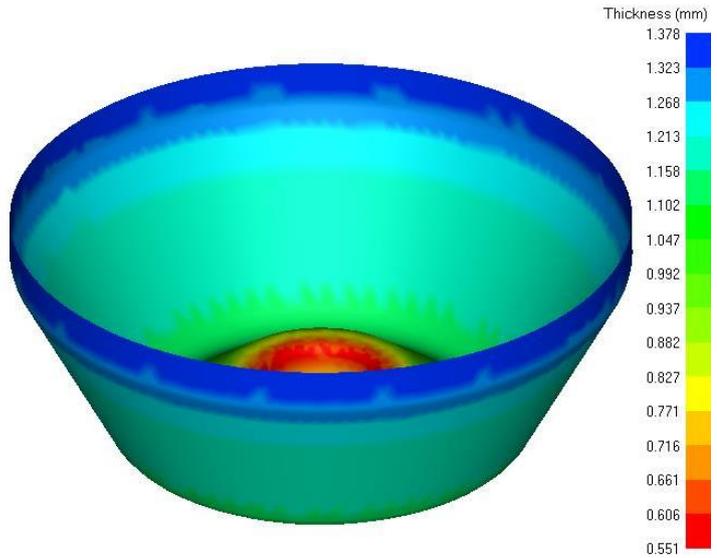

**Figure 5 (c) Thickness distribution of the modified component by DCI.**

The original component design and associated formulation for $Thickening$ is taken from Kakandikar 2014. The problem (Eq 17 to 19) was solved using SCI and DCI. The $Thickening$ (refer to Eq 17) depends on blank holder force, coefficient of friction, radius on die and radius on punch. All these parameters finally depend on variables blank diameter $d_o$, finished component diameter $d_1$ and corner radius $z$. These parameters define the design of the component. The solutions using the SCI and DCI are then used to modify the original component design.

It can be observed from the Table 7(a) that the thickness value obtained in the original component was $1.309\ mm$. The associated formability analysis performed in FromingSuite version 2015.1.0 software in (refer to Figure 5(a)) indicates that the range of the thickness distribution in the overall component varies from $0.356\ to\ 1.378\ mm$. The thickness obtained by SCI is $1.276\ mm$ (refer to Table 7(b)). It indicates that the thickness is reduced by $2.52\%$. Also it can be observed from Figure 5(b) that the whole component thickness distribution ranges from $0.357\ to\ 1.378\ mm$. It indicates that the average thickness through overall component is reduced. Similarly the thickness obtained by DCI is $1.268\ mm$ (refer to Table 7(c)). It indicates that the thickness is reduced by $3.13\%$. It can be observed from Figure 5(c) that the whole component thickness distribution ranges from $0.551\ to\ 1.378\ mm$, which indicates that the average thickness of the overall component is reduced.

## 5. Conclusion and Future Works

Two constraint handling approaches CI with static penalty approach (SCI) and CI with dynamic penalty approach (DCI) are successfully proposed and tested by solving 20 constrained test problems including pressure vessel design problem, tension-compression spring design and welded beam design problem. The results highlighted that the approach is significantly effective as compared to other algorithms solving these problems. The performance of the algorithm was satisfactory and competent in terms of the robustness, objective function value, and constraint satisfaction. The computational cost, i.e. computational time and function evaluations were also reasonable. In addition, the Springback effect problem occurring in the automotive punch plate, thinning in connector and thickening in tail cap during the process of deep drawing are successfully solved. This validated the applicability of the proposed constrained CI versions, SCI and DCI.

The computational time of CI can be further reduced by tuning the parameters such as size of cohort and sampling interval reduction factor. Also, the current version of SCI and DCI could solve problems with inequality constraints. An improvement in the existing approach is required to solve problems with equality constraints (Deshpande et al. 2013, Kulkarni & Tai 2011).

**Acknowledgements**

Authors would like to thank the anonymous reviewers. Their comments helped in much improvement in the quality of the manuscript.

**References**

Ali Kattan, Reem A. Alrawi, "Pressure Vessel Design Using the Dynamic Self-Adaptive Harmony Search Algorithm", SDIWC, ISBN: 978-0-9891305-4-7, pp.11-16, 2014.

Deshpande, A.M., Phatnani G.M., Kulkarni, A.J. (2013): "Constraint Handling in Firefly Algorithm", in Proceedings of IEEE International Conference on Cybernetics, Lausanne, Switzerland, 13-15 June 2013, pp. 186-190

Dorigo. M & Gambardella. L.M, "Ant Colony System: A Cooperative Learning Approach to the Travelling Salesman Problem", IEEE Transactions on Evolutionary Computation, vol. 1(1), pp. 53-66, 1997.

Eberhart R., Kennedy J., "A new optimizer using particle swarm theory.", Proceedings of the Sixth International Symposium on Micro Machine and Human Science, pp.39-43, 4-6 Oct 1995.

Efren Mezura-Montesa, Carlos A. Coello Coello," Constraint-handling in nature-inspired numerical optimization: Past, present and future", Swarm and Evolutionary Computation, vol. 1, pp. 173–194, 2011.

Efren Mezura-Montes, Carlos A. Coello Coello, "What Makes a Constrained Problem Difficult to Solve by an Evolutionary Algorithm", Technical Report, Evolutionary Computation Group, 2004.

Efren Mezura-Montes, Coello CAC, "A simple multimembered evolution strategy to solve constrained optimization problems", IEEE Trans Evol Computer 9(1):1–17 (2005).

Efren Mezura-Montes, Blanca Cecilia Lopez-Ramırez, "Comparing Bio-Inspired Algorithms in Constrained Optimization Problems", Journal of Evolutionary Computation, pp.662-669, 25-28, doi: 10.1109/CEC.2007.4424534, 2007.

Fister. I, Fister.I. Jr., Yang X.S, Brest J, "A comprehensive review of firefly algorithms", Journal of Swarm and Evolutionary Computation, vol.13, Issue.1, pp.34-46, 2013.

Homaifar A, S.H.Y. Lai, X. Qi, "Constrained optimization via genetic algorithms", Simulation 62 (4) (1994) 242–254.

Hong-Shuang Li, Siu-Kui Au, "Solving constrained optimization problems via Subset Simulation", 4th International Workshop on Reliable Engineering Computing, National University of Singapore, 2010.

Arora. J. "Introduction to optimum design", Academic Press, ISBN: 0-12-064155-0, 2004.

Jianjun Liu, Teo K. L., Xiangyu Wang, Changzhi Wu, "An exact penalty function-based differential search algorithm for constrained global optimization", Journal of Soft Computing, DOI 10.1007/s00500-015-1588-6, 2015.

Joines J, Houck C, "On the use of non-stationary penalty functions to solve nonlinear constrained optimization problems with Gas", in: D. Fogel (Ed.), Proceedings of the First IEEE Conference on Evolutionary Computation, IEEE Press, Orlando, FL, 1994, pp. 579–584.

Kakandikar G.M., Nandedkar V.M., "Optimization of forming load and variables in deep drawing process for automotive cup using Genetic Algorithm". IISc Centenary-International Conference on Advances in Mechanical Engineering ICICAME, Bangalore, 2008.

Kakandikar G. M., "Some studies on Optimization of Sheet Metal Forming Processes", Ph. D. Thesis, 2014, Shri Guru Gobind Singhji Institute of Engineering and Technology, Vishnupuri, Nanded 2014.

Karaboga Dervis, Bahriye Akay, "A modified Artificial Bee Colony (ABC) algorithm for constrained optimization problems", Applied Soft Computing, vol. 11, pp.3021–3031, 2011.

Krishnasamy G, Kulkarni A. J., Paramesran R, "A Hybrid Approach for Data Clustering Based on Modified Cohort Intelligence and K-means", Expert systems with applications, vol. 41, pp. 6009-6016, 2014.

Kulkarni, A.J., Baki, M.F., Chaouch, B.A.: "Application of the Cohort-Intelligence Optimization Method to Three Selected Combinatorial Optimization Problems", European Journal of Operational Research, 250(2), pp. 427-447, 2016.

Kulkarni A.J., Durugkar I.P., Kumar M.," Cohort intelligence: A self-supervised learning behavior", Proceedings of the IEEE International Conference on Systems, Man, and Cybernetics, pp.1396-1400, 2013.

Kulkarni, A.J., Krishnasamy, G., Abraham, A.: "Cohort Intelligence: A Socio-inspired Optimization Method", Intelligent Systems Reference Library, 114 (2017) Springer, (DOI 10.1007/978-3-319-44254-9), (ISBN: 978-3-319-44254-9)

Kulkarni A. J., Shabir Hinna, "Solving 0–1 Knapsack Problem using Cohort Intelligence Algorithm", International Journal of Machine Learning & Cybernetics, 7(3), pp. 427-441, 2016.

Kulkarni, A.J. and Tai, K. (2011) "Solving Constrained Optimization Problems Using Probability Collectives and a Penalty Function Approach", International Journal of Computational Intelligence and Applications, 10(4), pp. 445-470

Liang J. J., Runarsson T. P., Efren Mezura-Montes, Clerc Maurice, Suganthan P. N, Carlos A. Coello Coello, K. Deb, "Problem Defnitions and Evaluation Criteria", Special Session on Constrained Real-Parameter Optimization, September 18, 2006.

Lucidi S., Rinaldi F., "Exact penalty functions for nonlinear integer programming problems", J. Optimization Theory and Applications 145 (2010) 479–488.

Ma C, Li X., On an exact penalty function method for semi-infinite programming problems, Journal of Industrial and Management Optimization, 8(3), 2012.

Ma C, Zhang L. S. "On an exact penalty function method for nonlinear mixed discrete programming problems and its applications in search engine advertising problems", Applied Mathematics and Computations, 271, 2015.

Michalewicz Z, Attia N.F, "Evolutionary optimization of constrained problems", in: Proceedings of the 3rd Annual Conference on Evolutionary Programming, World Scientific, Singapore, 1994, pp. 98‑108.

Mirjalili Seyedali, Mirjalili S.M, Andrew Lewis, "Grey Wolf Optimizer", Advances in Engineering Software, Volume 69, Pages 46-61, ISSN 0965-9978, March 2014.

Mitchell M, "An Introduction to Genetic Algorithms", ISBN 0–262–13316–4 (HB), 0–262–63185–7 (PB), MIT press, Cambridge 1996.

Powell D, Skolnick M.M, "Using genetic algorithms in engineering design optimization with non-linear constraints", in: S. Forrest (Ed.), Proceedings of the Fifth International Conference on Genetic Algorithms, University of Illinois at Urbana-Champaign, Morgan Kaufmann, San Mateo, CA, July 1993, pp. 424‑431.

Qin A.K., Huang V.L., Suganthan P.N., "Differential evolution algorithm with strategy adaptation for global numerical optimization", IEEE Transactions on Evolutionary Computations, vol.13, issue. 2, pp. 398–417, 2009.

Schoenauer M, Michalewicz Z, "Evolutionary computation at the edge of feasibility", in: H.-M. Voigt, W. Ebeling, I. Rechenberg, H.-P. Schwefel (Eds.), Proceedings of the Fourth Conference on Parallel Problem Solving from Nature, Springer, Berlin, September 1996, pp. 245‑254.

Storn R., Price K, "Differential evolution–a simple and efficient heuristic for global optimization over continuous spaces.", Journal of Global Optimization, vol.11, issue. 4, pp.341–359, 1997.

Huyer. W., Neumair A, "A new exact penalty function", SIAM Journal of Optimization, 13 (2003) 1141–1159.

Yang X. S., Christian Huyck, Mehmet Karamanoglu, Nawaz Khan, "True Global Optimality of the Pressure Vessel Design Problem: A Benchmark for Bio-Inspired Optimization Algorithms", International Journal of Bio-Inspired Computation, vol. 5, issue. 6, pp. 329-335 2013.

Yang X.S., Deb S.,"Cuckoo search via Lévy flights.", World Congress on Nature & Biologically Inspired Computing, NaBIC, pp.210-214, 9-11 Dec. 2009.

Zavala AEM, Aguirre AH, Diharce ERV, "Constrained optimization via particle evolutionary swarm optimization algorithm (PESO)", In: Proceedings of the 2005 conference on genetic and evolutionary computation (GECCO'05), pp 209–216 (2005).